\documentclass[sigconf]{acmart}
\pdfoutput=1
\usepackage{booktabs} 
\usepackage{bm}
\usepackage{amsmath}
\usepackage{amssymb}
\usepackage{algorithm}
\usepackage{algorithmic}

\setcopyright{rightsretained}

\copyrightyear{2018} 
\acmYear{2018} 
\setcopyright{acmcopyright}
\acmConference[MM '18]{2018 ACM Multimedia Conference}{October 22--26, 2018}{Seoul, Republic of Korea}
\acmBooktitle{2018 ACM Multimedia Conference (MM '18), October 22--26, 2018, Seoul, Republic of Korea}
\acmPrice{15.00}
\acmDOI{10.1145/3240508.3240585}
\acmISBN{978-1-4503-5665-7/18/10}

\begin{document}
\title{CA$\bf{^3}$Net: Contextual-Attentional Attribute-Appearance Network \\ for Person Re-Identification}

\author{Jiawei Liu, Zheng-Jun Zha, Hongtao Xie, Zhiwei Xiong, Yongdong Zhang}

\affiliation{%
  \institution{University of Science and Technology of China}
}
\email{ljw368@mail.ustc.edu.cn}
\email{{zhazj, htxie, zwxiong, zhyd73}@ustc.edu.cn}

\thanks{Corresponding author: Zheng-Jun Zha (zhazj@ustc.edu.cn)}
%
%
%

\renewcommand{\shortauthors}{Jiawei Liu et al.}

\begin{abstract}
Person re-identification aims to identify the same pedestrian across non-overlapping camera views. Deep learning techniques have been applied for person re-identification recently, towards learning representation of pedestrian appearance. This paper presents a novel Contextual-Attentional Attribute-Appearance Network ($\rm CA^3Net$) for person re-identification. The $\rm CA^3Net$ simultaneously exploits the complementarity between semantic attributes and visual appearance, the semantic context among attributes, visual attention on attributes as well as spatial dependencies among body parts, leading to discriminative and robust pedestrian representation. Specifically, an attribute network within $\rm CA^3Net$ is designed with an Attention-LSTM module. It concentrates the network on latent image regions related to each attribute as well as exploits the semantic context among attributes by a LSTM module. An appearance network is developed to learn appearance features from the full body, horizontal and vertical body parts of pedestrians with spatial dependencies among body parts. The $\rm CA^3Net$ jointly learns the attribute and appearance features in a multi-task learning manner, generating comprehensive representation of pedestrians. Extensive experiments on two challenging benchmarks, i.e., Market-1501 and DukeMTMC-reID datasets, have demonstrated the effectiveness of the proposed approach.
\end{abstract}

%
%

\keywords{Person re-identification, attribute, appearance, deep learning}

\maketitle

\begin{figure}[!t]
	\begin{center}
		\includegraphics[width=1.0\linewidth]{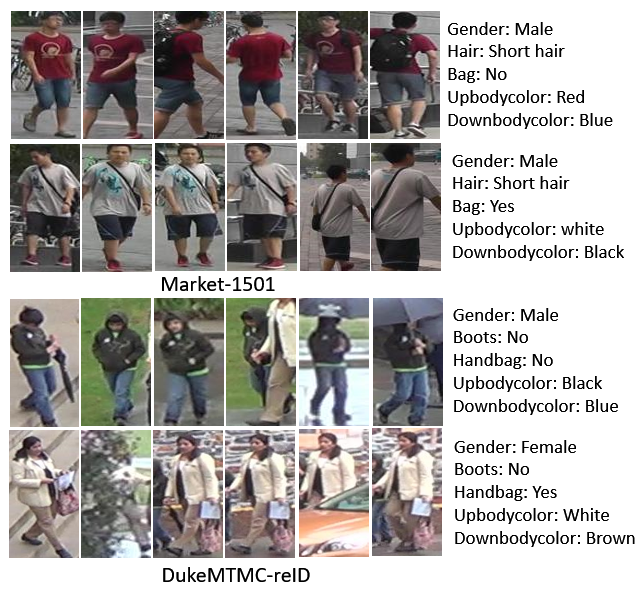}
	\end{center}
	\caption{Example pedestrian images with several corresponding attributes in the Market-1501 and DukeMTMC-reID person re-identification datasets, respectively.}
	\label{fig:long}
	\label{fig:onecol}
\end{figure}

\section{Introduction}
Person re-identification aims at identifying a target pedestrian at diverse locations over different non-overlapping camera views. It has attracted increasing attention recently because of its importance for many practical applications, such as automated surveillance, activity analysis and content-based visual retrieval \textit{etc} \cite{40,59}. Despite recent progress in person re-identification, it still remains a challenging task due to various challenges, including background clutter, occlusion, dramatic variations in illumination, body pose and viewpoint, as well as similar appearance among different pedestrians \textit{etc}. Figure 1 illustrates sample images of pedestrians with some attributes in two benchmarks of person re-identification, i.e., Market-1501 \cite{17} and DukeMTMC-reID \cite{18}.

Conventional person re-identification approaches are mainly based on hand-crafted descriptors of pedestrian appearance \cite{41,43}, such as Symmetry-Driven Accumulation of Local Features (SDALF) \cite{1}, Local Maximal Occurrence (LOMO)~\cite{2} and Weighted Histograms of Overlapping Stripes (WHOS) \cite{54} \textit{etc}. Recently, deep learning technique has been applied for person re-identification \cite{46,47,48,50}, towards learning discriminative appearance representation for identifying the same pedestrian and distinguishing different ones in an end-to-end manner. These approaches abstract global appearance features from full body of pedestrians, local appearance features from body parts or both of them by designing a variety of deep architectures. However, appearance representation, whether hand-crafted descriptors or deep learning features, is not robust to the aforementioned challenges, resulting in unsatisfactory person re-identification results.

On the other hand, person attributes, such as \textit{long hair}, \textit{short sleeve} and \textit{carrying a handbag} \textit{etc}, represent intermediate-level semantic properties of a pedestrian, providing crucial cues for identifying the pedestrian. Compared to appearance features, person attributes possess much better robustness to the variations of illumination, body pose and camera viewpoint \cite{12}. For pedestrians with similar appearance or a pedestrian having large appearance variance across images, appearance features usually result in false matches among pedestrians, As intermediate-level semantic descriptors, person attributes have shown good efficacy on dealing with such large intra-category variance and small inter-category variance within low-level feature space \cite{58,57}. Moreover, attribute and appearance representation of pedestrians are complementary to each other. They describe pedestrians from intermediate-level semantic abstraction and low-level visual details, respectively. The joint exploration of them could offer a comprehensive representation of pedestrians and thus enhance the accuracy of person re-identification. Recently, a few of preliminary works \cite{9,10,11} exploit attributes for person re-identification. They applied the attribute classifiers trained on auxiliary datasets to generate attribute responses over pedestrian images, which are in turn used for identifying pedestrians. However, they only used attribute representation and neglected visual description of appearance features that contain essential visual cues for re-identification. Lin \textit{et al.} \cite{12} annotated person attributes on the pedestrian images within the Market-1501 and DukeMTMC-reID person re-identification datasets. They also made a preliminary effort on learning attribute and appearance features for person re-identification. In \cite{12}, attributes are modeled individually without exploration of the semantic context among them. However, different attributes correlate semantically. Some attributes usually co-occur in a pedestrian image, while some ones are not likely to co-occur. Hence, the presence or absence of a certain attribute provides valuable cues for inferring the presence/absence of other related attributes. Moreover, an attribute usually arises from one or more regions within the images rather than entire images. Hence, it is essential to concentrate on the latent related image regions with visual attention mechanism when modeling attributes.

In this work, we propose a Contextual-Attentional Attribute-Appearance network ($\rm CA^3Net$) to learn discriminative and robust representation for person re-identification. The $\rm CA^3Net$ jointly learns attribute and appearance representation by multi-task learning for person identification and attribute recognition. $\rm CA^3Net$ simultaneously exploits the semantic context among attributes, visual attention on attributes, spatial context among body parts as well as the complementarity  between semantic attributes and visual appearance. As illustrated in Figure 2, $\rm CA^3Net$ consists of an attribute network learning attribute representation, an appearance network learning global and local appearance features as well as a base network generating low-level feature maps. Specifically, the attribute network contains an Attention-LSTM module, a convolution layer and a series of fully connected layers. The LSTM cell \cite{49} in the Attention-LSTM module captures the underlying semantic context among attributes which could effectively boost the learning of attributes. The attention block in the Attenion-LSTM module explores latent spatial attention for each attribute with identity-level supervision and concentrates the network's attention on the related image regions when learning attributes. The appearance network contains three convolution layers, three avg-pooling layers and ten fully connected layers. It learns appearance features from full body of pedestrians, horizontal body parts as well as vertical body parts with exploration of spatial dependencies among body parts along horizontal and vertical directions. The appearance network is trained by a part loss and a global loss which compute the identification errors on each body part and full body respectively to avoid over-fitting on a specific body part. The base network is built upon ResNet-50 \cite{13} model for extracting low-level visual details. Based on the above subnetworks, $\rm CA^3Net$ is able to learn effective representation of pedestrians leading to satisfactory person re-identification results. We conduct extensive experiments to evaluate $\rm CA^3Net$ on two widely-used person re-identification datasets, \textit{i.e.}, Market-1501 and DukeMTMC-reID, and report superior performance over state-of-the-art approaches.

The main contribution of this paper is three-fold: (1) We propose a novel Contextual-Attentional Attribute-Appearance Network ($\rm CA^3Net$) for person re-identification; (2) We design a new attribute learning network with an Attention-LSTM module, which exploits latent semantic context among attributes and visual attention on attributes; (3) We conduct extensive evaluations on two benchmark with significant performance improvements over state-of-the-art solutions.

\section{Related Work}

\begin{figure*}[!t]
	\centering
	\includegraphics[width=7in]{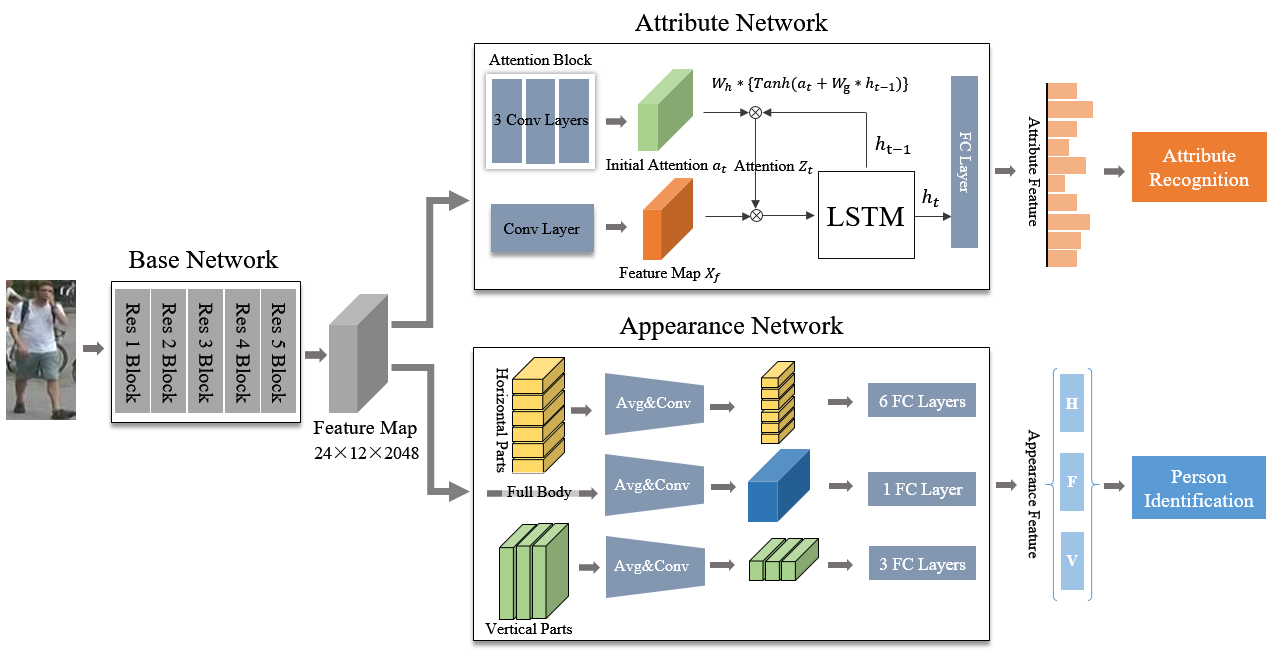}
	\caption{The overall architecture of the proposed $\rm CA^3Net$ approach, consisting of a base network, an attribute network and an appearance network.}
	\label{fig_overallss}
\end{figure*}

Recent years have witnessed many research efforts and encouraging progress on person re-identification.
This section briefly reviews existing works belonging to two major categories, i.e., appearance based re-identification and the newly emerging attribute based re-identification methods.

\textbf{Appearance based re-ID}. Appearance based person re-identifi-
cation approaches mainly focus on developing distinctive appearance representations from pedestrian images. Many sophisticated hand-crafted features have been developed to boost the performance. For example, Farenzena \textit{et al.} \cite{1} proposed the Symmetry-Driven Accumulation of Local Features (SDALF) to exploit the symmetry property of human body to handle variations of camera viewpoint. Liao \textit{et al.} \cite{2} analyzed the horizontal occurrence of local features, and proposed an effective feature called Local Maximal Occurrence (LOMO). Recently, deep learning technique has been adopted for person re-identification, towards learning discriminative representation of pedestrian appearance. For example, Liu \textit{et al.} \cite{40} proposed a multi-scale triplet CNN which captures visual appearance of a person at various scales by a comparative similarity loss on massive sample triplets. Li \textit{et al.} \cite{5} proposed to jointly learn global and local features with pre-defined grid horizontal stripes on pedestrian images by a multi-loss function. Li \textit{et al.} \cite{6} designed a Multi-Scale Context-Aware Network (MSCAN) to learn appearance features over full body and body parts with local context knowledge by stacking multi-scale convolutions, as well as a Spatial Transformer Networks (STN) to deal with the problem of pedestrian misalignment. Li \textit{et al.} \cite{7} formulated a Harmonious Attention CNN (HA-CNN) for the joint learning of soft pixel attention and hard region attention. Si \textit{et al.} \cite{8} proposed a Dual ATtention Match-ing network (DuATM) for learning context-aware feature sequences, in which both intra-sequence and inter-sequence attention strategies are used for feature refinement and feature-pair alignment, respectively.

\textbf{Attribute based re-ID}. Person attributes have been exploited for person re-identification in a few of recent works and have shown good robustness against challenging variations of illumination, body pose and viewpoint. Some of them adopt transfer learning technique to learn attributes for person re-identification task but neglect appearance feature. For example, Shi \textit{et al.} \cite{9} present a new semantic attribute learning model which was trained on a fashion photography dataset and adapted to provide a semantic description for person re-identification. Su \textit{et al.} \cite{10} proposed a weakly supervised multi-type attribute learning framework which involved a three-stage training to progressively boost the accuracy of attributes with only a limited number of labeled samples. Schumann \textit{et al.} \cite{11} developed a person re-identification approach which trained an attribute classifier on separate attribute dataset and integrated its responses into the person re-identification model based on CNNs. Recently, Lin \textit{et al.} \cite{12} annotated person attributes on the pedestrian images within the Market-1501 and DukeMTMC-reID datasets. They also proposed an attribute-person recognition (APR) network to learn attribute and appearance features for person re-identification. However, their work overlooks the semantic context among attributes and visual attention on attributes, which are importance for learning person attributes.
\section{The proposed Method}
In this section, we first present the overall architecture of the proposed $\rm CA^3Net$ and then elaborate its components.

\subsection{Overall Architecture}
Given a training set $\boldsymbol{X} = \{\boldsymbol{x}_i\}^{N}_{i=1}$ containing $N$ samples from $K$ pedestrians captured by non-overlapping camera networks together with their corresponding person ID as $\boldsymbol{Y}=\{\boldsymbol{y}_i\}_{i=1}^{N}$, the objective is to learn a discriminative representation for identifying the same pedestrian and distinguishing different pedestrians. We propose a novel Contextual-Attentional Attribute-Appearance Network ($\rm CA^3Net$), which simultaneously exploits complementarity between semantic attributes and visual appearance, semantic context among attributes, visual attention on attributes as well as spatial dependency among body parts. As shown in Figure 2, $\rm CA^3Net$ consists of an attribute network learning person attributes, an appearance network characterizing pedestrian appearance as well as a base network generating low-level visual representation. Specifically, the base network is built on ResNet-50 model \cite{13} due to its strong ability in learning visual representation. It consists of five Residual blocks, each of which contains several convolution layers with Batch Normalization (BN), Rectified Linear Units (ReLU) and optional Max-Pooling operations. The attribute network is proposed to effectively abstract attribute representation $f(\boldsymbol{x}_i)_{att}$ with an Attention-LSTM module. The attention block  concentrates the network on latent image patches that are related to each attribute during attribute learning. The LSTM cell progressively takes each attribute as input and decides whether to retain or discard the latent semantic context from the current attribute and previous ones. It leverages the inter-attribute context for attribute recognition. Furthermore, we develop an appearance network to learn appearance feature $f(\boldsymbol{x}_i)_{app}$ from the full body, horizontal body parts and vertical body parts of pedestrians simultaneously. The body-part features are learned with a part loss function leading to detailed visual cues from each body part, while the full-body features are learned with a global loss function extracting the global visual appearance of pedestrians.

The attributes and appearance features are learned with the loss functions for person re-identification and attribute recognition, respectively, in a multi-task learning manner. During testing, the two features are integrated to form the final pedestrian representation. The matching score between pedestrian images $\boldsymbol{x}_i$ and $\boldsymbol{x}_j$ can be computed as:
\begin{equation}
\begin{split}
S(\boldsymbol{x}_i,\boldsymbol{x}_j) = \big\|[f(\boldsymbol{x}_i)_{app};f(\boldsymbol{x}_i)_{att}]-[f(\boldsymbol{x}_j)_{app};f(\boldsymbol{x}_j)_{att}]\big\|^2_2
\end{split}
\end{equation}


\subsection{Attribute Network}
An attribute learning network is proposed to learn discriminative intermediate-level semantic descriptions of pedestrian images by the task of attribute recognition. An attribute usually arises from one or more regions within the images. The network is expected to concentrate on the corresponding regions when learning an attribute. However, such regions are not localized with ground-truth. On the other hand, different attributes correlate semantically. The presence or absence of a certain attribute is usually useful for inferring the presence/absence of other related attributes. For example, the attributes ``wearing a dress'' and ``long hair'' are likely to co-occur, the attributes ``carrying a bag'' and ``carrying a backpack'' may mutually exclusive. The exploration of such semantic context among attributes can well boost attribute recognition. Motivated by these observations, we propose a novel attribute network with an Attention-LSTM module, which contains an attention block and a LSTM cell. The attention block consisting of 3 convolution layers, learning spatial attention for each attribute. The LSTM cell sweeps all attributes sequentially, memorizes the semantic correlation and dependencies from previous inputs by the memory mechanism. As shown in Figure 2, the attribute network consists of a convolution layer, an Attention-LSTM module and $C$ fully-connected layers. $C$ is the number of attributes.

The attention block, containing 3 consecutive convolution layers, takes the feature tensor $\boldsymbol{T}$ as input to generate initial attention maps $\boldsymbol{A}$ for all attributes. $\boldsymbol{T}$ is the output feature maps of the base network. Each channel in the initial attention maps $\boldsymbol{A}$ ($24\times12\times C$) corresponds to one attribute. The kernel sizes of the 3 convolution layers are $1\times1\times512$,  $3\times3\times256$, and $1\times1\times C$, respectively. The BN and ReLU nonlinearity operations are performed with the first two convolution layers. In addition, another individual convolution layer with kernel size $1\times1\times256$ is utilized to transfer $\boldsymbol{T}$ to feature map $\boldsymbol{X}_f$. The attributes are regarded as a temporal sequence. At each time step $t$, the LSTM cell receives an attentional attribute feature map $\boldsymbol{x}_t$ corresponding to attribute $C_t$ as input, which comes from the element-wise multiplication of $\boldsymbol{X}_f$ and the precise attention map $\boldsymbol{Z}_t$, and outputs attribute predictions for the attribute recognition task. The feedback connections and internal gating mechanism of the LSTM cell is able to memorize the latent semantic dependencies among attributes, selectively discover and propagate relevant context to next attribute. The formulation of the LSTM cell is shown as follows
\begin{equation}
\begin{split}
\left( \begin{array}{c}
\bm{i}_{t}\\
\bm{f}_{t}\\
\bm{o}_{t}\\
\bm{g}_{t}
\end{array} \right) &= \left( \begin{array}{c}
\sigma\\
\sigma\\
\sigma\\
tanh
\end{array} \right)\bm{M}\left(\begin{array}{c}
\bm{x}_t\\
\bm{h}_{t-1}
\end{array}\right)\\
\bm{c}_t &= \bm{f}_t\odot \bm{c}_{t-1} + \bm{i}_t\odot \bm{g}_{t}\\
\bm{h}_t &= \bm{o}_t\odot tanh(\bm{c}_t)
\end{split}
\end{equation}
where $\boldsymbol{f}_t$, $\boldsymbol{i}_t$, $\boldsymbol{o}_t$, $\boldsymbol{c}_t$, $\boldsymbol{M}$, and $\boldsymbol{h}_t$ are the forget gate, input gate, output gate, cell state, weight matrix and hidden state respectively. In addition, $\boldsymbol{X}_{in} = (x_1,x_2,...x_t,...x_C)$ represents the set of attentional attribute feature maps for all attributes. The LSTM cell sweeps all person attributes and generates discriminative attribute features.

In order to learn more precise spatial distribution of attention over a pedestrian image for each attribute, we incorporate the initial attention map $\boldsymbol{A}_t$ and previous internal hidden state $\boldsymbol{h}_{t-1}$ to obtain a precise attention map $\boldsymbol{Z}_t$ of attribute $C_t$, instead of using $\boldsymbol{A}_t$ for attribute $C_t$ directly. The formulations of the incorporation of $\boldsymbol{A}_t$ and $\boldsymbol{h}_{t-1}$ is shown as follows:

\begin{equation}
\boldsymbol{U}_t = \boldsymbol{W}_h*(F_{tanh}(\boldsymbol{a}_t + \boldsymbol{W}_g*\boldsymbol{h}_{t-1}))
\end{equation}
where $\boldsymbol{W}_h \in \mathbb{R}^{k\times k}$ and $\boldsymbol{W}_g \in \mathbb{R}^{k\times d}$ are parameters to be learned, $k$ is the area of the feature Tensor $\boldsymbol{T}$ ($24\times12$), $d$ is the dimension of the hidden state (256). $\boldsymbol{U}_t$ is referred to the unnormalized attention map. Then the precise attention map $Z_t$ can be obtained by spatial normalization with a softmax function

\begin{equation}
Z^{i,j}_t = \frac{exp(U^{i,j}_t)}{\sum_{i,j}exp(U^{i,j}_t)}, \quad \mathbf{Z}\in\mathbb{R}^{24\times12\times C}
\end{equation}
where $Z^{i,j}_t$ denotes the normalized attention values at the pixel $(i,j)$ for attribute $C_t$. Figure 3 illustrates the attention maps corresponding to certain attributes, such as shoes, hat, upper-body clothing and handbag. It can be observed that the regions corresponding to the attributes are get concentrated with high attention scores. This indicates that the attention block enables the network to concentrate on the regions corresponding to attributes and thus generate more precise modeling of attributes. Finally, each type of attribute feature is taken into a m-dim FC layer and a corresponding softmax function layer, where $m$ represents the specific attribute has $m$ categories. The attribute network utilizes the Attention-LSTM module to capture visual attention for each attribute and explore the semantic context, which is effective for improving the performance of attribute recognition.

\begin{figure}[htb]
	\centering
	\includegraphics[width=3.2in]{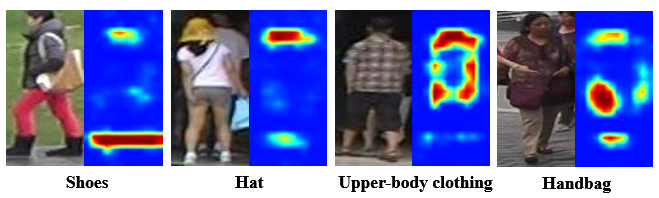}
	\caption{Visualization of the attention maps corresponding to certain attributes for some pedestrian images.}
	\label{fig_overallss}
\end{figure}

\subsection{Appearance Network}
An appearance network is developed to learn global and local appearance representation by the task of person re-identification.  Existing methods usually abstract appearance feature from full body of pedestrians and/or from horizontal body parts. For learning more effective appearance feature for person matching, an appearance network is designed to extract representation from full body of pedestrians and horizontal body parts, as well as the vertical body parts with exploration of spatial dependencies among body parts along both horizontal and vertical directions.

As shown in Figure 2, the appearance network consists of 3 convolution layers, 3 average-pooling layers and 10 fully-connected (FC) layers. Each convolution layer is followed with a BatchNorm (BN) layer and a Rectified Linear Units (ReLU) layer. The appearance network takes the feature tensor $\boldsymbol{T}$ as input, and partitions $\boldsymbol{T}$ into $h$ horizontal stripes and $v$ vertical stripes of body, respectively. Then, the feature tensor $\boldsymbol{T}$, the horizontal and vertical stripes go through three corresponding local branches to abstract global appearance feature and part-based features, respectively. We define the vector of activations along the channel axis as a column vector. The three local branches in the network employ mean-pooling layers to average all the column vectors in a same horizontal stripe or a same vertical stripe for producing a horizontal part-level column vector $\boldsymbol{g}_i$  $(i=1,2,...h)$ or a vertical part-level column vector $\boldsymbol{k}_i$ $(i=1,2,...v)$, as well as average $\boldsymbol{T}$ to produce a global-level column vector $\boldsymbol{p}$. Afterward, three convolution layers are applied to reduce the dimension of the three type column vectors, respectively. Finally, each type column vector is taken into a classifier layer which is implemented with a FC layer and a corresponding softmax function layer for classifying the person ID.

The size of $\boldsymbol{T}$ is $24\times12\times2048$, which is also equally partitioned into 6 horizontal stripes ($4\times12\times2048$) and 3 vertical stripes ($24\times4\times2048$), respectively. The kernel sizes of the three convolution layers in the network are $1\times1\times256$. The dimension of the FC layers is the total number of person IDs. The appearance network is optimized by minimizing the sum of part loss for $h$ horizontal stripes and $v$ vertical stripes, and global loss for $\boldsymbol{T}$. During testing stage, all the pieces of 256-dimensional column vectors are concatenated to form the appearance feature $f(\boldsymbol{x}_i)_{app} = [\boldsymbol{g}_1, \boldsymbol{g}_2,..., \boldsymbol{g}_h, \boldsymbol{k}_1, \boldsymbol{k}_2,..., \boldsymbol{k}_v, \boldsymbol{p}]$, which is used for pedestrian matching.

\subsection{Loss Function and Optimization}
Identification loss is usually leveraged for classification task and has advantages in terms of simplicity and effectiveness.  Hence, we adopt the identification loss to optimize the appearance features and attribute features. Suppose the training set $\boldsymbol{D}_i = \{x_i, y_i, c_i\}$ has $N$ images of $K$ identities, where $x_i$ denotes the $i$-$th$ person images, $y_i$ is person IDs of image $x_i$ and $c_i = \{c_i^1, c_i^2, ..., c_i^L\}$ is a set of $L$ attribute labels of person image $x_i$. Given training examples, the proposed $\rm CA^3Net$ extracts $\boldsymbol{x}_i^h$, $\boldsymbol{x}_i^v$ (local appearance features) and  $\boldsymbol{x}_i^g$ (global appearance feature) from the appearance network, as well as extract $L$ attribute features $\boldsymbol{x}_i^a$ from the attribute network. The loss function for the task of person re-identification is the sum of part loss and global loss, which is formulated as follows:

\begin{equation}
\begin{split}
\mathcal{L}_{app} &= \sum\limits_{i=1}^{H}\mathcal{L}(\boldsymbol{x}_i^h) + \sum\limits_{i=1}^{V}\mathcal{L}(\boldsymbol{x}_i^v) + \mathcal{L}(\boldsymbol{x}_i^g)\\
\mathcal{L}(x_i)& = -\frac{1}{S}\sum\limits_{i=1}^{S}\log{\frac{e^{\boldsymbol{W}^T_{y_i}\boldsymbol{x}_i+\boldsymbol{b}}}{\sum_{j=1}^Ke^{\boldsymbol{W}^T_j\boldsymbol{x}_j+\boldsymbol{b}}}}
\end{split}
\end{equation}
where $y_i$ is the corresponding person ID of the $i$-th pedestrian image, $\boldsymbol{W}_j\in \mathbb{R}^D$ represents the j-th column of the weight matrix $\boldsymbol{W}\in\mathbb{R}^{D\times K}$ and $\boldsymbol{b}$ refers to a bias term. $H$ and $V$ are the number of horizontal stripes and vertical stripes, respectively. The batch size of the input is $S$.

The loss function for the attribute classification task is the sum of $L$ attribute classification losses, which is formulated as follows:
\begin{equation}
\mathcal{L}_{att}=\sum\limits_{i=1}^{L}\mathcal{L}(x_i^a)
\end{equation}

By jointing the person re-identification task and attribute classification task, the proposed $\rm CA^3Net$ is optimized to predict person IDs and attributes, simultaneously. The total loss function for the $\rm CA^3Net$ is defined as follows:
\begin{equation}
\mathcal{L} = \mathcal{L}_{app} + \lambda\cdot \mathcal{L}_{att}
\end{equation}
where $\lambda$ denotes the balance weight of the two loss functions.


\section{Experiments}
In this section, we conduct extensive experiments to evaluate the performance of the proposed $\rm CA^3Net$ on two widely used person re-identification datasets and compare the $\rm CA^3Net$ to state-of-the-art methods. The experimental results show that $\rm CA^3Net$ achieves superior performance of person re-identification over the state-of-the-art methods. Moreover, we investigate the effectiveness of the proposed $\rm CA^3Net$ including the attribute network and the appearance network.

\textbf{Datasets} - There are several benchmark datasets established for person re-identification. In this work, extensive experiments are conducted on two widely used datasets, \textit{i.e}, Market-1501 and DukeMTM
C-reID for fair comparison and evaluation. The two person re-identification datasets are challenging and realistic.

The Market-1501 dataset is one of the largest and most realistic person re-identification benchmark, contains 32,643 images of 1,501 identities captured by 6 cameras. All images are automatically detected by the Deformable Part Model (DPM) detector \cite{15}. Following the protocol used in \cite{17}, the dataset is fixedly divided into two parts respectively, one part contains 12,936 images of 750 identities as training set and the other contains 19,732 images of 751 identities as testing set. The proposed method is compared to the state-of-the-art methods under single query evaluation setting.

The DukeMTMC-reID dataset is a subset of the DukeMTMC dataset \cite{19} and is one of the most challenging re-ID datasets due to similar clothes of different pedestrians and occlusion by trees and cars. It contains 36,411 hand-drawn bounding boxes of 1,812 identities from 8 high-resolution cameras. Following the evaluation protocol specified in \cite{18}, it is fixedly divided into two parts respectively, one part contains 16,522 images of 702 identities as training set and the other contains 17,661 gallery images of 702 identities as testing set. In addition, there are 2,228 query pedestrian images. Analogously, performance on the DukeMTMC-reID dataset is also evaluated under single query evaluation setting.

Pedestrian images in the Market-1501 dataset are annotated with 27 attributes at identity-level. Each attribute is labeled with its presence or absence on each pedestrian image, the attribute ``age'' is labeled with four types, \textit{i.e.}, young, teenager, adult and old. Considering that there are 8 and 9 colors (\textit{eg},  upblack, upwhite and downred \textit{etc}) for upper-body clothing and lower-body clothing respectively and only one color is labeled as positive for one identity, we regard the 8 upper-body colors and the 9 lower-body colors as one upper-body clothing color attribute with 9 classes and one lower-body clothing color attribute with 10 classes (there is one more category for the case that the upper-body clothing or lower-body clothing colors of a pedestrian may not belong to the 8 or 9 colors). Hence, there are 12 attributes consisting of binary and multi-type valued attributes on the Market-1501 dataset as well as 10 attributes in the DukeMTMC-reID dataset. As the Attention-LSTM module processes attributes sequentially, we need to determine the order of attributes. However, person attributes are naturally without a fixed order. A promising solution is to adopt multiple orders of attributes (e.g., rare first, frequent first, top-down, and random order \textit{etc.}) and fuse their results for subsequent module \cite{56}. In the experiments, we adopt two types of orders, i.e., top-down according to body topological structure and fine-abstract following to the semantic granularity from fine grained attributes to abstract attributes.

\textbf{Implementation Details} - The implementation of the proposed method is based on the Pytorch framework with two NVID-
IA Titan XP GPUs. We adopt the pre-trained model on ImageNet to initialize parameters of the $\rm CA^3Net$ on the two person re-identificat-
ion datasets. The stochastic gradient descent (SGD) algorithm is started with learning rate $lr$ of 0.01, the weight decay of $5e^{-4}$ and the Nesterov momentum of 0.9. The parameter $\lambda$ in Eq. (7) is set to 2. All the images are resized to the size of $384\times192\times3$ and normalised with $1.0/256$. Meanwhile, the training set is enlarged by data augmentation strategies \cite{20} including random horizontal flipping and random erasing probability of 0.5 during training phase. The number of mini-batches is set to 64. The proposed network is optimized for 250 iterations in each epoch, and 70 epochs in total. Moreover, the whole training process is divided into three parts. In the first stage, the base network followed with the appearance network is trained until convergence for person re-identification task, which impels the base network to learn befitting feature maps prepared for the attribute classification task. In the second stage, the $\rm CA^3Net$ is trained for person re-identification and attribute classification tasks and learn discriminative appearance feature and robust attribute feature, respectively. In the last stage, the two features are merged and the $\rm CA^3Net$ is re-trained for only person re-identification task until convergence, which can drive the $\rm CA^3Net$ to learn discriminative and robust representation of pedestrian for person re-identification.

\textbf{Protocol} - Cumulative Matching Characteristic (CMC) is extensively adopted for quantitative evaluation of person re-identification methods. The rank-$k$ recognition rate in the CMC curve indicates the probability that a query identity appears in the top-$k$ position. The other evaluation metric is the mean average precision (mAP), considering person re-identification as a retrieval task.

\subsection{Comparison to State-of-the-Arts}

\begin{table}[htbp]
	\begin{center}
		\newcommand{\tabincell}[2]{\begin{tabular}{@{}#1@{}}#2\end{tabular}}
		\begin{tabular}{|c|c|c|c|c|}
			\hline
			\textbf{Method}&\textbf{Rank-1}&\textbf{Rank-5}&\textbf{Rank-10}&\textbf{mAP}\\
			\hline
			Bow+kissMe\cite{17}&44.4&63.9&72.2&20.8\\
			\hline
			WARCA\cite{24}&45.2&68.1&76.0&-\\
			\hline
			KLFDA\cite{23}&46.5&71.1&79.9&-\\
			\hline
			DNS\cite{22}&55.43&-&-&29.9\\
			\hline
			CRAFT\cite{21}&68.7&87.1&90.8&42.3\\
			\hline
			\hline
			SOMAnet\cite{25}&73.9&88.0&92.2&47.9\\
			\hline
			HydraPlus\cite{27}&76.9&91.3&94.5&-\\
			\hline
			SVDNet\cite{26}&82.3&92.3&95.2&62.1\\
			\hline
			PAN\cite{28}&82.8&-&-&63.4\\
			\hline
			Triplet Loss\cite{29}&84.9&94.2&-&69.1\\
			\hline
			MultiScale \cite{30}&88.9&-&-&73.1\\
			\hline
			GLAD \cite{31}&89.9&-&-&73.9\\
			\hline
			HA-CNN\cite{7}&91.2&-&-&75.7\\
			\hline
			\hline
			ACRN \cite{11}&83.6&92.6&95.3&62.6\\
			\hline
			APR\cite{12}&84.3&93.2&95.2&64.7\\
			\hline
			\hline
			$\rm CA^3Net$&\textbf{93.2}&\textbf{97.8}&\textbf{98.6}&\textbf{80.0}\\
			\hline $\rm CA^3Net$(RK)&\textbf{94.7}&\textbf{97.0}&\textbf{97.8}&\textbf{91.5}\\
			\hline
		\end{tabular}
	\end{center}
	\caption{Performance comparison to the state-of-the-art methods on the Market-1501 dataset.}
\end{table}

\begin{table}[htbp]
	\begin{center}
		\newcommand{\tabincell}[2]{\begin{tabular}{@{}#1@{}}#2\end{tabular}}
		\begin{tabular}{|c|c|c|c|c|}
			\hline
			\textbf{Method}&\textbf{Rank-1}&\textbf{Rank-5}&\textbf{Rank-10}&\textbf{mAP}\\
			\hline
			Bow+kissMe\cite{17}&25.14&-&-&12.2\\
			\hline
			LOMO+XQDA\cite{2}&30.8&-&-&17.0\\
			\hline
			\hline
			GAN\cite{18}&67.7&-&-&47.1\\
			\hline
			PAN\cite{28}&71.6&83.9&-&45.0\\
			\hline
			SVDNet\cite{26}&76.7&86.4&89.9&56.8\\
			\hline
			MultiScale \cite{30}&79.2&-&-&60.6\\
			\hline
			EMR\cite{36}&80.4&-&-&63.9\\
			\hline
			HA-CNN\cite{7}&80.5&-&-&63.8\\
			\hline
			Deep-Person\cite{37}&80.9&-&-&64.8\\
			\hline
			\hline
			APR\cite{12}&70.7&-&-&51.9\\
			\hline
			ACRN\cite{11}&72.6&84.8&88.9&52.0\\
			\hline
			\hline
			$\rm CA^3Net$&\textbf{84.6}&\textbf{92.4}&\textbf{94.9}&\textbf{70.2}\\
			\hline
			$\rm CA^3Net$(RK)&\textbf{89.6}&\textbf{94.1}&\textbf{95.4}&\textbf{86.4}\\
			\hline
		\end{tabular}
	\end{center}
	\caption{Performance comparison to the state-of-the-art methods on the DukeMTMC-reID dataset.}
\end{table}

\textbf{Market-1501:} Table 1 shows the performance comparison of the proposed $\rm CA^3Net$ against 15 state-of-the-art methods in terms of CMC accuracy and mAP. The compared methods belong to three categories, \textit{i.e.}, traditional methods based on hand-crafted feature and/or distance metric learning including Bow+kissMe \cite{17}, WARCA \cite{24}, KLFDA \cite{23}, DNS \cite{22} and CRAFT \cite{21}, deep learning based methods including SOMANet \cite{25}, HydraPlus \cite{27}, SVDNet \cite{26}, PAN \cite{28}, Triplet Loss \cite{29}, MultiScale \cite{30}, GLAD \cite{31} and HA-CNN \cite{7} and attribute based methods including ACRN \cite{11} and APR \cite{12}. The proposed $\rm CA^3Net$ achieves 93.2\% rank-1 recognition rate and 80.0\% mAP score. We can see that our method surpasses existing methods, improving the 2nd best compared method HA-CNN by 2.2\% rank-1 recognition rate and 5.7\% mAP score. Moreover, $\rm CA^3Net$ achieves significant performance improvement compared to the two attribute based methods ACRN and APR by 11.5\% and 10.6\% at rank-1 recognition rate, respectively. The comparison indicates that the effectiveness of the proposed $\rm CA^3Net$ for jointly exploiting attributes and appearance information. Moreover, the Attention-LSTM in the attribute network is able to capture latent semantic context among attribute and learn the spatial attention for each attribute. $\rm CA^3Net$(RK) refers to the proposed method with re-ranking \cite{38} with k-reciprocal encoding, which is an effective strategy for boosting the performance. With the help of re-ranking, rank-1 accuracy and mAP of $\rm CA^3Net$ are further improved to 94.7\% and 91.5\% respectively.

\textbf{DukeMTMC-reID:} We compare the proposed $\rm CA^3Net$ against 11 state-of-the-art methods including two traditional methods Bow
+kissMe \cite{17} and LOMO+XQAD \cite{24}, and seven deep learning based methods including GAN \cite{18}, PAN \cite{28}, SVDNet \cite{26}, MultiScale \cite{30}, EMR \cite{36}, HA-CNN \cite{7} and Deep-Person \cite{37}, as well as two attribute based methods APR \cite{12} and  ACRN \cite{11}. From Table 2, we can observe that the proposed $\rm CA^3Net$ outperforms all the existing methods at all ranks, obtaining the best 84.6\% rank-1 recognition rate and 70.2\% mAP score. $\rm CA^3Net$ boosts the 2nd best compared method Deep-Person by 4.6\% rank-1 recognition rate and 8.3\% mAP score. In addition, the performance of $\rm CA^3Net$ achieves 19.7\% and 16.5\% improvement of rank-1 accuracy respectively, compared to the attribute based methods APR and ACRN. The result of $\rm CA^3Net$ can be increased to 89.6\% rank-1 recognition rate and 86.4\% mAP score with re-Ranking. An illustration of some retrieval results is given in Figure 4.


\begin{figure*}[!t]
	\centering
	\includegraphics[width=6.3in]{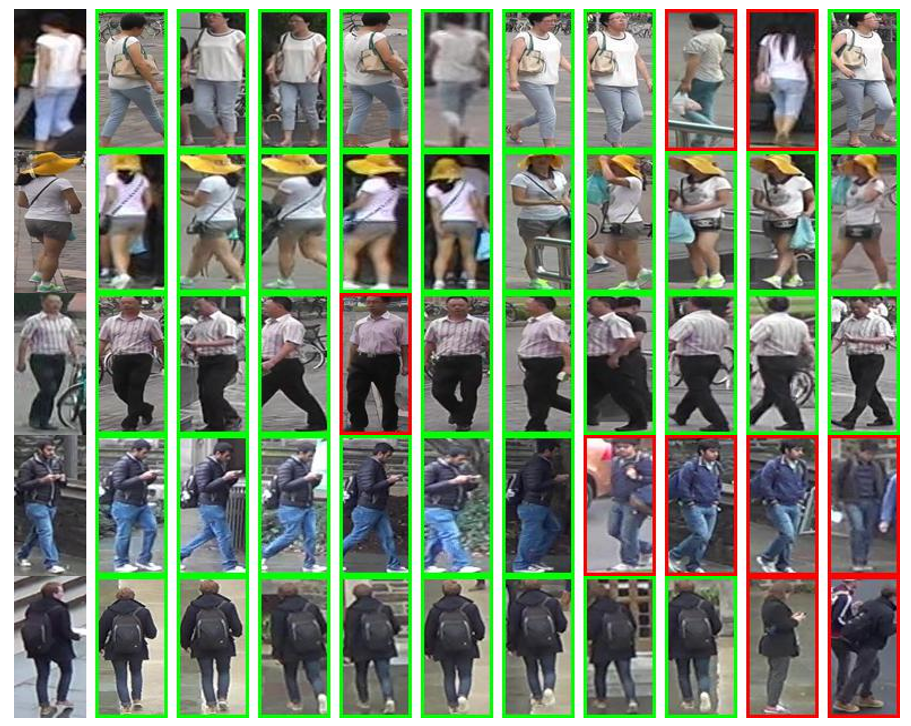}
	\caption{Example results of the proposed \textbf{$\rm CA^3Net$} on the Market-1501 (top three rows) and DukeMTMC-reID
		(bottom two rows) datasets. The images in the first column are queries. Correct matches are highlighted
		green and false matches in red.}
	\label{fig_overallss}
\end{figure*}

\subsection{Ablation Studies}
To demonstrate the effectiveness and contribution of each component of the $\rm CA^3Net$, we conduct a series of ablation experiments on the DukeMTMC-reID dataset. Moreover, we compare the performance of the different type of  appearance features for the appearance network. We also evaluate the effect of the Attention-LSTM  module for the attribute network.

Table 3 summarizes the ablation results of the proposed $\rm CA^3Net$. $\rm CA^3Net$\_w/o App refers to $\rm CA^3Net$ without the appearance network, which exploits the base network followed with the attribute network to learn attribute feature and use the attribute features to match pedestrians directly.  $\rm CA^3Net$\_w/o Att refers to $\rm CA^3Net$ without the attribute network, which utilizes the base network followed with the appearance network to learn appearance feature. From Table 3, we can observe that $\rm CA^3Net$\_w/o App only obtains 57.1 \% rank-1 accuracy, since the network is not specially designed for person re-identification task. $\rm CA^3Net$\_w/o Att achieves performance of 80.1 \% rank-1 accuracy. $\rm CA^3Net$ yields the best performance of 84.6\% than the other two networks, which shows the effectiveness of $\rm CA^3Net$ for joint exploration of both appearance and attribute features, leading to satisfactory person re-identification results.

\begin{table}[htbp]
	\large
	\begin{center}
		\newcommand{\tabincell}[2]{\begin{tabular}{@{}#1@{}}#2\end{tabular}}
		\begin{tabular}{|c|c|c|}
			\hline
			\textbf{Model}&\textbf{Rank-1}&\textbf{mAP}\\
			\hline
			$\rm CA^3Net$\_w/o App&57.1&33.3\\
			\hline
			$\rm CA^3Net$\_w/o Att&80.1&64.1\\
			\hline
			$\rm CA^3Net$&84.6&70.2\\
			\hline
			
		\end{tabular}
	\end{center}
	\caption{Evaluation of the effectiveness of each component within $\rm CA^3Net$ on the DukeMTMC-reID dataset.}
\end{table}

\begin{table}[htbp]
	\large
	\begin{center}
		\newcommand{\tabincell}[2]{\begin{tabular}{@{}#1@{}}#2\end{tabular}}
		\begin{tabular}{|c|c|c|}
			\hline
			\textbf{Model}&\textbf{Rank-1}&\textbf{mAP}\\
			\hline
			AppNet\_G&72.1&51.9\\
			\hline
			AppNet\_V&77.6&59.6\\
			\hline
			AppNet\_H&79.2&63.9\\
			\hline
			AppNet&80.1&64.1\\
			\hline
		\end{tabular}
	\end{center}
	\caption{Evaluation of the effectiveness of the global and local appearance features on the DukeMTMC-reID dataset.}
\end{table}

\begin{table}[htbp]
	\large
	\begin{center}
		\newcommand{\tabincell}[2]{\begin{tabular}{@{}#1@{}}#2\end{tabular}}
		\begin{tabular}{|c|c|c|}
			\hline
			\textbf{Model}&\textbf{Rank-1}&\textbf{mAP}\\
			\hline
			AttNet\_Base&40.3&21.8\\
			\hline
			AttNet\_LSTM&43.9&22.8\\
			\hline
			AttNet\_Attention&53.7&31.2\\
			\hline
			AttNet&57.1&33.3\\
			\hline
		\end{tabular}
	\end{center}
	\caption{Evaluation of the effectiveness of each component within Attention-LSTM module on the DukeMTMC-reID dataset.}
\end{table}

Table 4 reports the accuracy of different type of appearance features. All the experiments are conducted with the base network followed by the appearance network. AppNet\_G only extracts global appearance feature for person re-identification, achieving 72.1\% rank-1 accuracy and 51.9\% mAP score. AppNet\_V only abstracts local appearance feature from vertical stripes of body, obtaining 77.6 \% rank-1 accuracy and 59.6\% mAP score. AppNet\_H is to extract local appearance feature from horizontal stripes of body, acquiring 79.2\% rank-1 accuracy and 63.9\% mAP score. AppNet extracts both the global and local appearance features and obtains the best performance of 80.1\% rank-1 accuracy and 64.1\% mAP score. By comparing different type of appearance features, it can be observed that the performance of local appearance feature surpasses the global appearance feature, which indicates that the local feature is able to guides the network to learn more detailed visual cues and avoid over-fitting on a specific body part. Moreover, joint learning global and local features enforces the appearance network to learn more spatial context among different body parts and are effective to improve the performance of person re-identification.

Table 5 compares each component of Attention-LSTM module in the attribute network. All the experiments are conducted with the base network followed by the attribute network. AttNet\_Base dose not use Attention-LSTM module and directly connects the output feature tensor \boldmath$T$\unboldmath with each attribute classifier, obtaining 40.3\% rank-1 accuracy. AttNet\_LSTM is to replace the Attention-LSTM module with a single LSTM cell, obtaining 43.9 \% rank-1 accuracy. AttNet\_Attention only uses the attention block of the full Attention-LSTM module, achieving 53.7\% rank-1 accuracy. AttNet is to employ the Attention-LSTM module for the attribute network and get the best performance of 57.1\% rank-1 accuracy. By comparing the results of AppNet\_Base and AppNet\_LSTM, the LSTM cell is able to learn the contextual semantic correlation of attributes. Meanwhile, from the comparison of AppNet\_Base and AppNet\_Attention, the attention block is able to learn a precise spatial attention for each attribute. Therefore, the Attention-LSTM module can improve the accuracy of attribute recognition and learn more discriminative attribute feature for person re-identification.

\section{Conclusions}
In this work, we proposed a novel Contextual-Attentional Attribute-Appearance Network ($\rm CA^3Net$) to learn discriminative and robust pedestrian representation for person re-identification. The proposed $\rm CA^3Net$ jointly learns semantic attribute and visual appearance representation of pedestrians with simultaneous exploration of the semantic context among attributes, visual attention on attributes and spatial dependencies between body parts. The exploration of inter-attribute context and visual attention leads to precise learning of attributes, in turn generating effective attribute representation. The appearance features learned from both full body and body parts provide a comprehensive description of pedestrian appearance. We conducted extensive experiments on two widely-used real-world person re-identification datasets, i.e., Market-1501 and DukeMTMC-reID. The experimental results have shown that the proposed $\rm CA^3Net$  achieves significant performance improvements over a wide range of state-of-the art methods.

\begin{acks}
This work was supported by the National Natural Science Foundation of China (NSFC) under Grants 61622211, 61472392, 61620106009 and 61525206 as well as the Fundamental Research Funds for the Central Universities under Grant WK2100100030. 
\end{acks}


\bibliographystyle{ACM-Reference-Format}
\bibliography{sample-bibliography}

\end{document}